\documentclass{llncs}
\usepackage{times}

\usepackage{makeidx}  % allows for indexgeneration
\usepackage{graphicx}
\usepackage{amsmath,amssymb} % define this before the line numbering.
\usepackage{tikz}
\usepackage{float}
\usepackage{pgfplots}
\pgfplotsset{compat=newest}
\usepackage[hidelinks]{hyperref}

\newcommand{\showodsf}[3]{%
\mbox{\input{data/pr/#1_#2_#3_ods_f.txt}\hspace{-2.5pt}}%
}

\newcommand{\deflinewidth}{1.5pt} % Line width

\newcommand{\cfbox}[2]{%
    \colorlet{currentcolor}{.}%
    {\color{#1}%
    \fbox{\color{currentcolor}#2}}%
}

\begin{document}

\mainmatter              % start of the contributions
\title{Deep Retinal Image Understanding}
%
%\titlerunning{Hamiltonian Mechanics}  % abbreviated title (for running head)
\author{Kevis-Kokitsi Maninis\inst{1}, Jordi Pont-Tuset\inst{1}, Pablo Arbel\'{a}ez\inst{2}, and Luc Van Gool\inst{1,3}}
% index{Maninis    , Kevis-Kokitsi}
% index{Pont-Tuset , Jordi        }
% index{Arbel\'{a}ez, Pablo        }
% index{Van Gool   , Luc          }
\institute{$^1$ETH Z\"urich $\qquad$ $^2$Universidad de los Andes $\qquad$ $^3$KU Leuven}
%\institute{Princeton University, Princeton NJ 08544, USA,\\
%\email{I.Ekeland@princeton.edu}%,\\ WWW home page:
%\texttt{http://users/\homedir iekeland/web/welcome.html}
%\and
%Universit\'{e} de Paris-Sud,
%Laboratoire d'Analyse Num\'{e}rique, B\^{a}timent 425,\\
%F-91405 Orsay Cedex, France}

\maketitle              % typeset the title of the contribution

\begin{abstract}
This paper presents Deep Retinal Image Understanding (DRIU), a unified framework of retinal image analysis that 
provides both retinal vessel and optic disc segmentation. We make use of deep Convolutional Neural Networks (CNNs), which have proven revolutionary in other fields of computer vision such as object detection and image classification, and we bring their power to the study of eye fundus images. DRIU uses a base network architecture on which two set of specialized layers are trained to solve both the retinal vessel and optic disc segmentation. We present experimental validation, both qualitative and quantitative, in four public datasets for these tasks. In all of them, DRIU presents super-human performance, that is, it shows results more consistent with a gold standard than a second human annotator used as control.

\end{abstract}
\section{Introduction}

Retinal image understanding is key for ophthalmologists while assessing widely spread eye diseases such as glaucoma, diabetic retinopathy, macular degeneration, and hypertension, among others. Although these diseases can lead to severe visual impairment and blindness if left untreated, early diagnosis and appropriate treatment, coupled with periodic examination by specialists, have proven to be determinant factors for controlling their evolution, which translates into better prognosis and an improved quality of life for patients. Given that several risk factors associated to these diseases, such as sedentarism, obesity and aging, are related to lifestyle, their frequency among the general population is growing. This situation has stressed the need for automated methods to assist ophthalmologists in retinal image understanding, and has sparked the interest for this field among the medical image analysis community.

Two anatomical structures are of particular interest for specialists when performing diagnostic measurements on eye fundus images: the blood vessel network and the optic disc. Consequently, most prior work on automated analysis of retinal images has focused on the segmentation of these two structures. Classic methods for addressing the task of blood vessel segmentation involve hand crafted filters like line detectors~\cite{RiPe07,Ngu+13} and vessel enhancement techniques~\cite{Soa+06,Zha+10,Fra+12a}. Approaches that rely on powerful machine learning techniques have emerged over the last years. In~\cite{OrBl14} the authors combine different kinds of vessel features and segment them with fully connected conditional random fields. In~\cite{Bec+13} a gradient boosting framework is proposed for learning filters in a supervised way. Algorithms that are able to enhance fine structures given a regression of retinal vessels have also been developed recently~\cite{SLF15,GuCh15}. Prior work on optic disc segmentation includes morphological operators~\cite{WaKl01} and hand crafted features~\cite{You+08} Morales \textit{et al.}~\cite{Mor+13} use morphology along with Principal Component Analysis (PCA) to obtain the structure of the optic disk inside a retinal image. A superpixel classification method is proposed in~\cite{Che+13}. 

In the last five years, deep learning techniques have revolutionized the field of computer vision. Deep Convolutional Neural Network (CNN) architectures were initially designed for the task of natural image classification~\cite{Krizhevsky2012}, and recent research has led to impressive progress in solving that problem. At the core of these approaches, lies a \emph{base network} architecture, starting from the seminal AlexNet~\cite{Krizhevsky2012}, to the more complex and more accurate VGGNet~\cite{SiZi15} and the inception architecture of GoogLeNet~\cite{Sze+15}. Furthermore, CNNs have been applied successfully to a large variety of general recognition tasks such as object detection~\cite{girshick2016region}, semantic segmentation~\cite{hariharan2015hypercolumns}, and contour detection~\cite{XiTu15}. In the domain of retinal image understanding, CNNs have been used for retinal vessel segmentation in~\cite{GaLe14} to classify patch features into different vessel classes. For optic disc segmentation, the authors of~\cite{Zil+15} use a CNN to extract features, which they post-process to obtain binary segmentations. Instead, our work is based on applying a CNN end-to-end, both for retinal vessel segmentation and optic disc detection, efficiently, since we avoid the redundant computations from a patch-based approach. 

\begin{figure}[t!]
\begin{minipage}[b]{\linewidth}
\centering
	\includegraphics[width=0.99\linewidth]{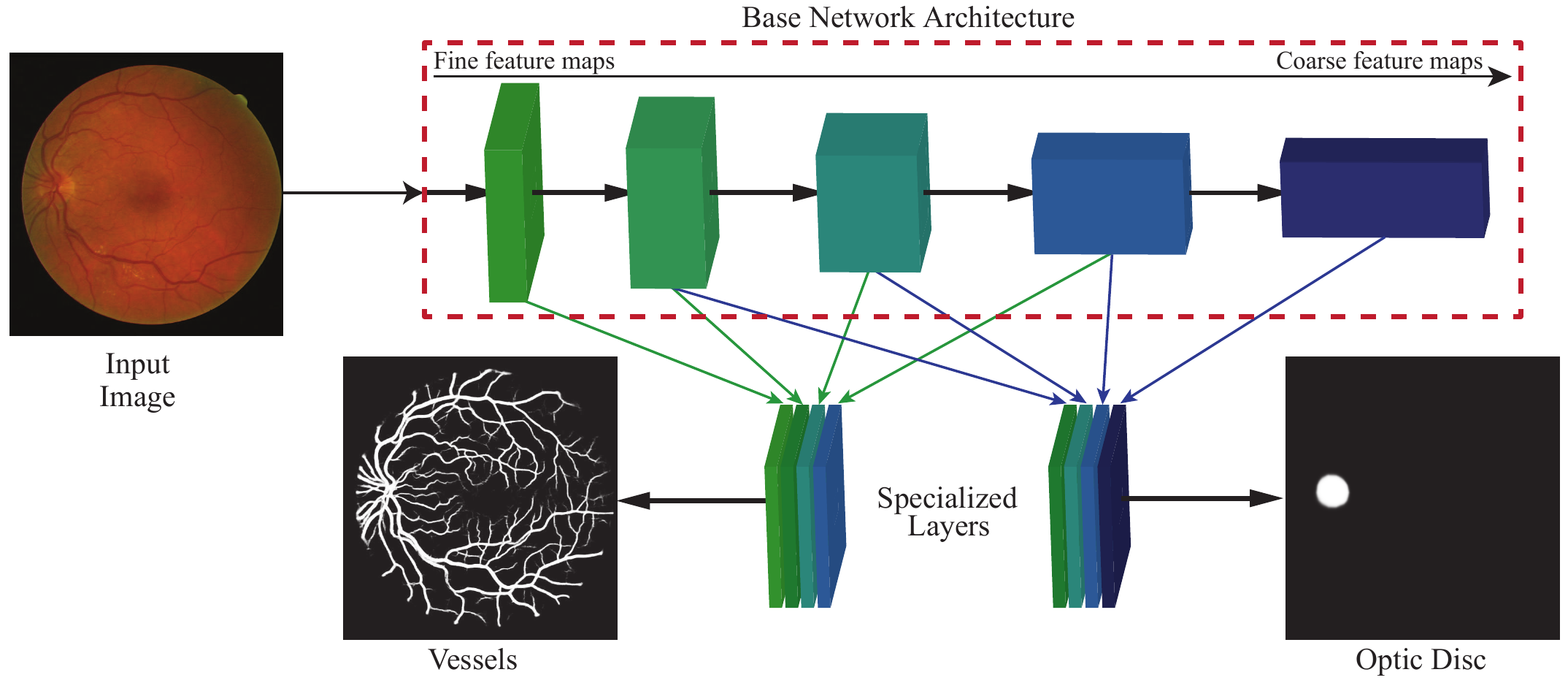}
\end{minipage}
\vspace{-7mm}
\caption{\textbf{Overview of DRIU}: Given a base CNN, we extract side feature maps and 
design specialized layers to perform blood vessel segmentation (left) and optic disc segmentation (right).}
\label{fig:DRIU}
\end{figure}

In this paper, we leverage the latest progress on deep learning to advance the field of automated retinal image interpretation. We design a CNN architecture that specializes a base network for the tasks of segmenting blood vessels and optic discs in fundus images. An overview of our approach, which we call Deep Retinal Image Understanding (DRIU), is presented in Figure~\ref{fig:DRIU}. DRIU is both highly efficient and highly accurate: at inference time, it requires a single forward pass of the CNN to segment both the vessel network and the optic disc and, as the experimental results will show, DRIU reaches or surpasses the performance of trained human specialists for both tasks on four publicly available annotated datasets.

\section{CNNs for Retinal Image Understanding}

We approach retinal vessel segmentation and optic disc detection as an image-to-image regression task, by designing a novel CNN architecture. We start from the VGG~\cite{SiZi15} network, originally designed for large-scale natural image classification. For our purposes, the fully connected layers at the end of the network are removed, so it mainly consists of convolutional layers coupled with Rectified Linear Unit (ReLU) activations. The use of four max pooling layers in the architecture separates the network into five stages (as in Figure~\ref{fig:DRIU}), each stage consisting of several convolutional layers. Between the pooling layers, feature maps of the same stage that are generated by convolutions with different filters have the same size. As we proceed deeper in the network, the information becomes coarser due to the decrease in size, which is a key factor for generalization. We call this part of our architecture the ``base network''. The layers of the base network are already pre-trained on millions of images, which is necessary for training deep architectures. To effectively use the information from feature maps with different sizes, we draw inspiration from the ``inception'' architecture of GoogLeNet~\cite{Sze+15}, which adds supervision at multiple internal layers of the network, and we connect task-specific ``specialized'' convolutional layers to the final layer of each stage. Each specialized layer produces feature maps in $K$ different channels, which are resized to the original image size and concatenated, creating a volume of fine-to-coarse feature maps. We append one last convolutional layer which linearly combines the feature maps from the volume created by the specialized layers into a regressed result. In our experiments, we used $K=16$. The majority of convolutional layers employ $3\times 3$ convolutional filters for efficiency, except the ones used for linearly combining the outputs ($1 \times 1$ filters).

 For training the network, we adopt the class-balancing cross entropy loss function originally proposed in~\cite{XiTu15} for the task of contour detection in natural images. We denote the training dataset by $S=\lbrace \left( X_n, Y_n \right),n=1,...,N \rbrace$, with $X_n$ being the input image and $Y_n=\lbrace y_j^{(n)},j=1,...,|X_{n}| \rbrace ,  y_j^{(n)} \in \lbrace 0,1 \rbrace$  the predicted pixel-wise labels. For simplicity, we drop the subscript $n$. The loss function is then defined as: \vspace{-10pt}

\footnotesize
\begin{align}
\mathcal{L}\left( \mathbf{W}\right)=-\beta\sum_{j	\in Y_+}{\log{P\left(y_j=1 |X;\mathbf{W}\right)}}-(1-\beta)\sum_{j	\in Y_-}{\log{P\left(\!y_j=0 |X;\mathbf{W}\right)}} \label{eq:hed_cost_side}
\end{align}
\normalsize
where $\mathbf{W}$ denotes the standard set of parameters of the CNN, which are trained with backpropagation. The multiplier $\beta$ is used to handle the imbalance of the substantially greater number of background compared to foreground pixels, which in our case are the vessels or the optic disc. Class balancing is necessary when we have severely biased ground truths (e.g.: approximately 10\% of the pixels are vessels, while the others are background). $Y_+$ and $Y_-$ denote the foreground and background sets of the ground truth $Y$, respectively. In this case, we use $\beta=|Y_-|/|Y|$. The probability $P \left( . \right)$ is obtained by applying a sigmoid $\sigma \left( . \right)$ to the activation of the final convolutional layer.

We use the same network architecture for both retinal vessel segmentation and optic disc segmentation. We found that the coarse feature maps of the final stage do not help with vessel detection since they contain coarse information which erases the thin vessels, whereas the finest ones of the first stage do not help with detecting the coarse structure of the optic disc. Thus, we construct two separate feature map volumes, one for each task. For retinal vessel segmentation, the volume contains features from the 4 finer stages, while for optic disc detection we use the 4 coarser stages (see Figure~\ref{fig:DRIU}). Our final result for both tasks is a probability map, in which a pixel detected as vessel/disc is assigned a higher score. 

At training time, we fine-tune the entire architecture (base network and specialized layers) for 20000 iterations. We use stochastic gradient descent with momentum, operating on one image per iteration. Due to the lack of data, the learning rate is set to a very small number ($lr=10^{-8}$), which is gradually decreased as the training process proceeds. We augment the datasets using standard techniques, by rotating and scaling the images, as a pre-processing step. We also substract the mean value of the training images for each colour channel.

At testing time, there is no need for any pre-processing of the data. The entire architecture runs on a GPU, and operates on the original RGB channels of a retinal image. The average execution time for retinal vessel and optic disc segmentation on an NVIDIA TITAN-X GPU is 85 milliseconds (ms) for DRIVE and 104 ms for STARE, which is orders of magnitude faster than the current state-of-the-art~\cite{OrBl14,Bec+13,GaLe14}. The same applies for optic disc segmentation, where our algorithm processes an image of DRIONS-DB in 65 ms and 110 ms for the larger images of RIM-ONE dataset.

\section{Experimental Validation}

We experiment on eye fundus images to segment both the blood vessels and the optic disc . In both cases, and for each database, we split the data into separate \textit{training} and \textit{test} sets, we learn all the parameters of DRIU on the training set, and then evaluate the final model on the previously unseen test set. Since CNNs are high-capacity models, it is a standard practice to augment the training set by rotating and scaling the images, in order to avoid learning regularities caused by the limited size of the dataset, such as the location of the optic disc in the training images, which would lead to overfitting and poor generalization. Additionally, we keep the learning rate of the network very low (fine-tuning with $lr=10^{-8}$).

\paragraph{\textbf{Blood Vessel Segmentation:}}
We experiment on the DRIVE~\cite{Sta+04} and STARE~\cite{HKG00} datasets (40 and 20 images, respectively). Both contain manual segmentations of the blood vessels by two expert annotators. We use the segmentations of the first annotator as the gold standard to train/test our algorithm. The ones from the second are evaluated against the gold standard to measure human performance.
For DRIVE we use the standard train/test split and for STARE we use the split defined in~\cite{GuCh15}, according to which the first 10 images consist the training set and the last 10 the test set.
We compare DRIU with the current state-of-the-art~\cite{GaLe14,Bec+13,OrBl14} as well as some traditional approaches for retinal vessel segmentation~\cite{RiPe07,Soa+06}. We also retrain HED~\cite{XiTu15} (state of the art in generic contour detection with CNNs) on DRIVE and STARE using their public code. 

\begin{figure}[h!]
\begin{minipage}[b]{0.48\linewidth}
\centering
%\scalebox{0.6}{%
\centering
	\includegraphics[width=0.325\linewidth]{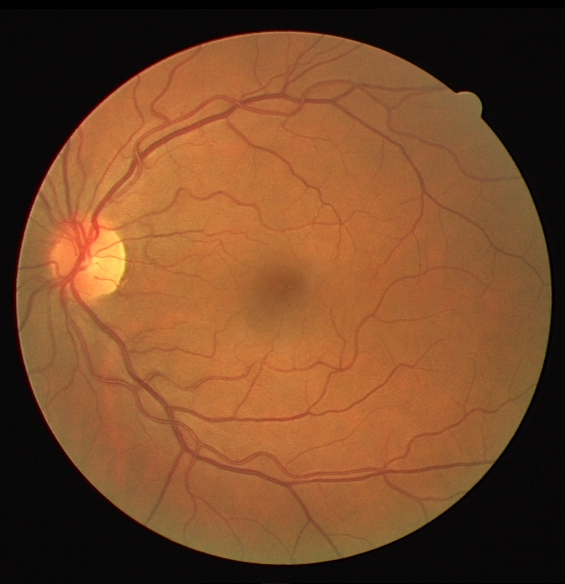}\hfill
	\includegraphics[width=0.325\linewidth]{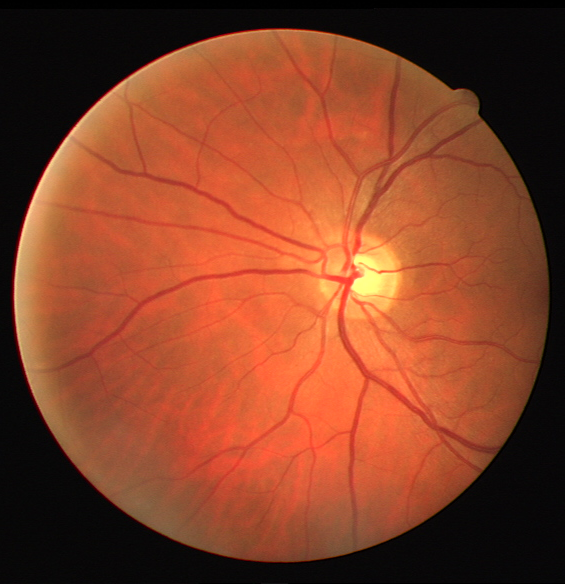}\hfill
	\includegraphics[width=0.325\linewidth]{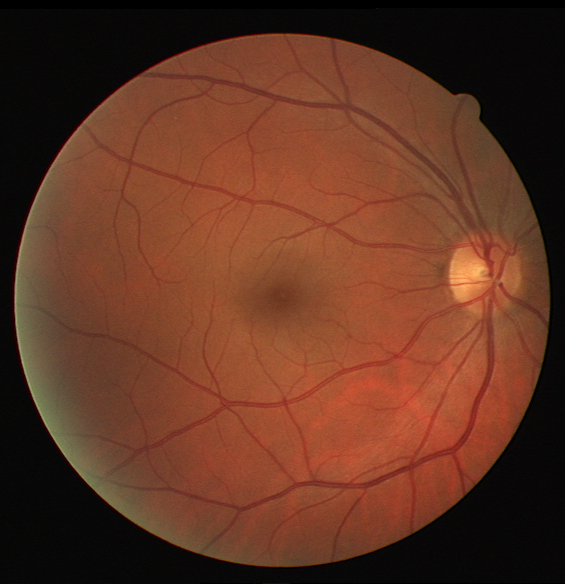}\\[1pt]
	\includegraphics[width=0.325\linewidth]{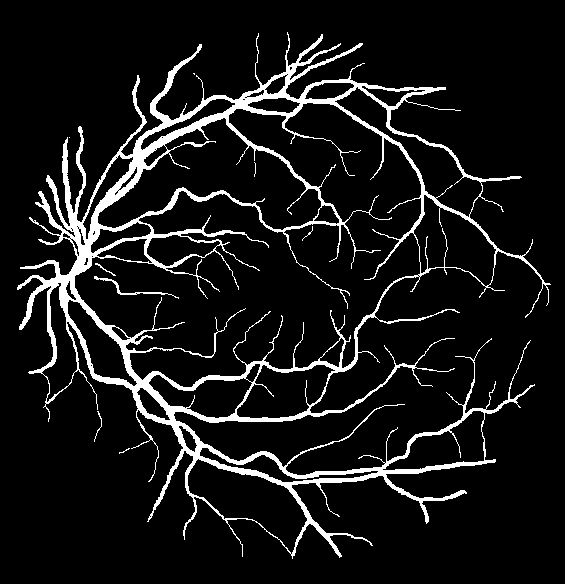}\hfill 
	\includegraphics[width=0.325\linewidth]{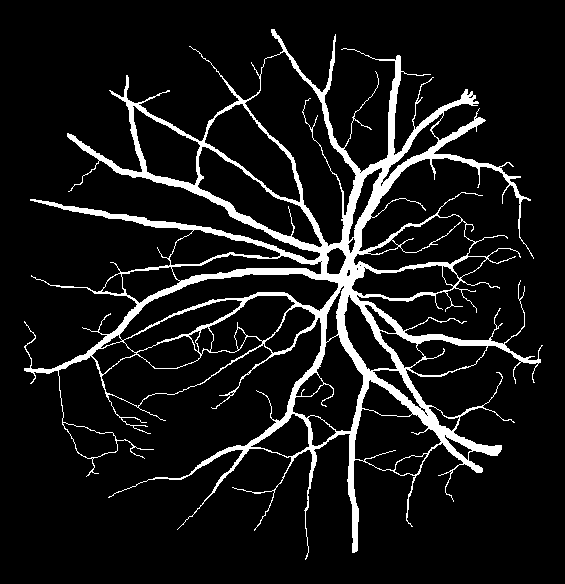}\hfill
	\includegraphics[width=0.325\linewidth]{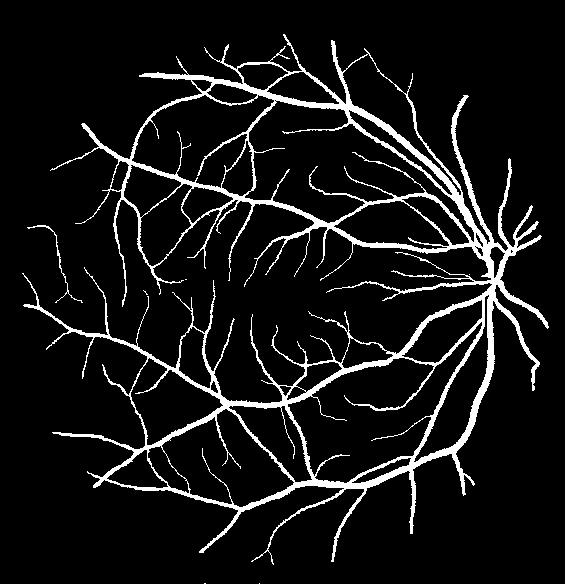}\\[1pt]
	\includegraphics[width=0.325\linewidth]{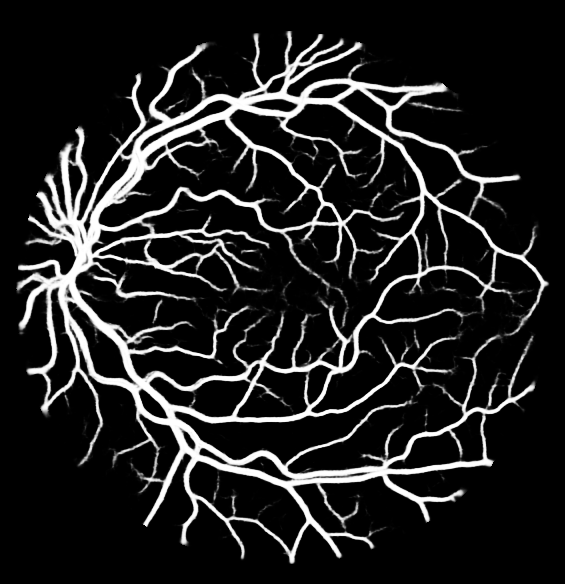}\hfill 
	\includegraphics[width=0.325\linewidth]{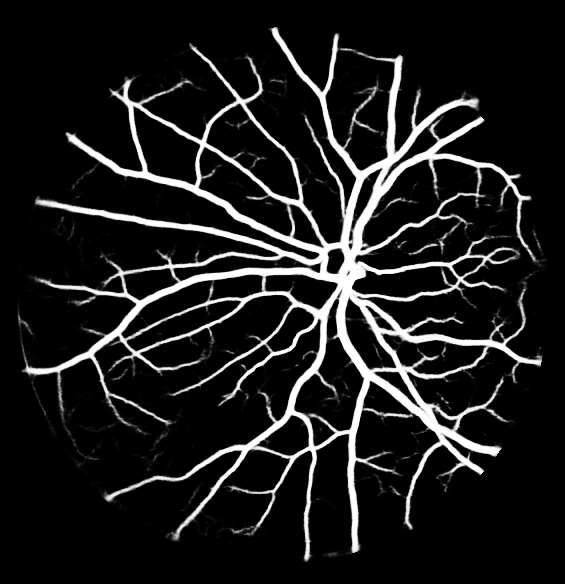}\hfill 
	\includegraphics[width=0.325\linewidth]{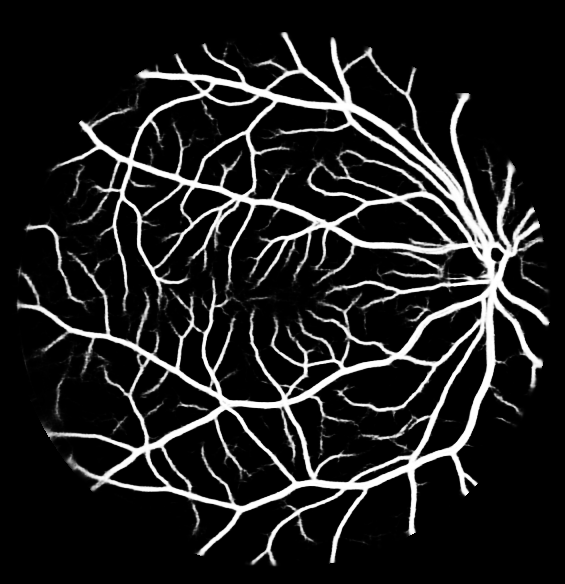} 	
\end{minipage}
\hfill
\begin{minipage}[b]{0.5\linewidth}
\centering
\scalebox{0.63}{%
\begin{tikzpicture}[/pgfplots/width=1.7\linewidth, /pgfplots/height=1.7\linewidth]
    \begin{axis}[% Axis labels
                 ymin=0.3,ymax=1,xmin=0.3,xmax=1,
    			 % Axis labels
        		 xlabel=Recall,
        		 ylabel=Precision,
         		 xlabel shift={-2pt},
        		 ylabel shift={-3pt},
         		 % General appearance
		         font=\small,
		         axis equal image=true,
		         enlargelimits=false,
		         clip=true,
		         % Grids 
        	     grid style=dotted, grid=both,
                 major grid style={white!65!black},
        		 minor grid style={white!85!black},
		 		 xtick={0,0.1,...,1.1},
        		 ytick={0,0.1,...,1.1},
         		 minor xtick={0,0.02,...,1},
		         minor ytick={0,0.02,...,1},
		         xticklabels={0,.1,.2,.3,.4,.5,.6,.7,.8,.9,1},
		         yticklabels={0,.1,.2,.3,.4,.5,.6,.7,.8,.9,1},
        		 % Legend
				 legend cell align=left,
        		 legend style={at={(0.02,0.02)},anchor=south west},
		         % Title
		         title style={yshift=-1ex,},
		         title={DRIVE - Region Precision Recall}]
        
    % Iso-f curves
    \foreach \f in {0.1,0.2,...,0.9}{%
       \addplot[white!50!green,line width=0.2pt,domain=(\f/(2-\f)):1,samples=200,forget plot]{(\f*x)/(2*x-\f)};
    }

	% Ours
    \addplot+[black,solid,mark=none, line width=\deflinewidth,forget plot] table[x=Recall,y=Precision] {data/pr/DRIVE_test_fb_Ours.txt};
    \addplot+[black,solid,mark=o, mark size=1.3, mark options={solid},line width=\deflinewidth] table[x=Recall,y=Precision] {data/pr/DRIVE_test_fb_Ours_ods.txt};
    \addlegendentry{[\showodsf{DRIVE_test}{fb}{Ours}] Ours}

    % N4_fields
    \addplot+[olive,solid,mark=none, line width=\deflinewidth,forget plot] table[x=Recall,y=Precision] {data/pr/DRIVE_test_fb_N4_fields.txt};
    \addplot+[olive,solid,mark=o, mark size=1.3, mark options={solid},line width=\deflinewidth] table[x=Recall,y=Precision] {data/pr/DRIVE_test_fb_N4_fields_ods.txt};
    \addlegendentry{[\showodsf{DRIVE_test}{fb}{N4_fields}] \textit{$N^4$fields}~\cite{GaLe14}}
    
	% Kernel_Boost
    \addplot+[red,solid,mark=none, line width=\deflinewidth,forget plot] table[x=Recall,y=Precision] {data/pr/DRIVE_test_fb_Kernel_Boost.txt};
    \addplot+[red,solid,mark=o, mark size=1.3, mark options={solid},line width=\deflinewidth] table[x=Recall,y=Precision] {data/pr/DRIVE_test_fb_Kernel_Boost_ods.txt};
    \addlegendentry{[\showodsf{DRIVE_test}{fb}{Kernel_Boost}] Kernel Boost~\cite{Bec+13}}

    	% HED
    \addplot+[green,solid,mark=none, line width=\deflinewidth,forget plot] table[x=Recall,y=Precision] {data/pr/DRIVE_test_fb_HED.txt};
    \addplot+[green,solid,mark=square, mark size=1.25, line width=\deflinewidth] table[x=Recall,y=Precision] {data/pr/DRIVE_test_fb_HED_ods.txt};
    \addlegendentry{[\showodsf{DRIVE_test}{fb}{HED}] HED~\cite{XiTu15}}
      
    % CRFs
    \addplot+[only marks,magenta,mark=asterisk,mark size=2.7,thick] table[x=Recall,y=Precision] {data/pr/DRIVE_test_fb_CRFs_ods.txt};
    \addlegendentry{[\showodsf{DRIVE_test}{fb}{CRFs}] \textit{CRFs}~\cite{OrBl14}}
    
    % Wavelets
    \addplot+[gray,solid,mark=none, line width=\deflinewidth,forget plot] table[x=Recall,y=Precision] {data/pr/DRIVE_test_fb_Wavelets.txt};
    \addplot+[gray,solid,mark=square, mark size=1.25, line width=\deflinewidth] table[x=Recall,y=Precision] {data/pr/DRIVE_test_fb_Wavelets_ods.txt};
    \addlegendentry{[\showodsf{DRIVE_test}{fb}{Wavelets}] Wavelets~\cite{Soa+06}}

    % Line Detectors
    \addplot+[orange,solid,mark=none,line width=\deflinewidth,forget plot] table[x=Recall,y=Precision] {data/pr/DRIVE_test_fb_Ricci_Line.txt};
    \addplot+[orange,solid,mark=x, mark size=1.6, mark options={solid},line width=\deflinewidth] table[x=Recall,y=Precision] {data/pr/DRIVE_test_fb_Ricci_Line_ods.txt};
    \addlegendentry{[\showodsf{DRIVE_test}{fb}{Ricci_Line}] Line Detectors~\cite{RiPe07}}
      
    % SE
    \addplot+[blue,solid,mark=none,line width=\deflinewidth,forget plot] table[x=Recall,y=Precision] {data/pr/DRIVE_test_fb_SE.txt};
    \addplot+[blue,solid,mark=x, mark size=1.6, mark options={solid},line width=\deflinewidth] table[x=Recall,y=Precision] {data/pr/DRIVE_test_fb_SE_ods.txt};
    \addlegendentry{[\showodsf{DRIVE_test}{fb}{SE}] \textit{SE}~\cite{DoZi13}}
    
    % Human
    \addplot+[only marks,blue,mark=+,mark size=2.7,ultra thick] table[x=Recall,y=Precision] {data/pr/DRIVE_test_fb_Human_ods.txt};
    \addlegendentry{[\showodsf{DRIVE_test}{fb}{Human}] \textit{Human}}
    
    \end{axis}
 \end{tikzpicture}}
 \vspace{-19pt}
\end{minipage}\\[2mm]
\hrule\rule{0mm}{2mm}\\
\begin{minipage}[b]{0.48\linewidth}
\centering
%\scalebox{0.6}{%
\centering
\setlength{\fboxsep}{0pt}
	\includegraphics[width=0.36\linewidth]{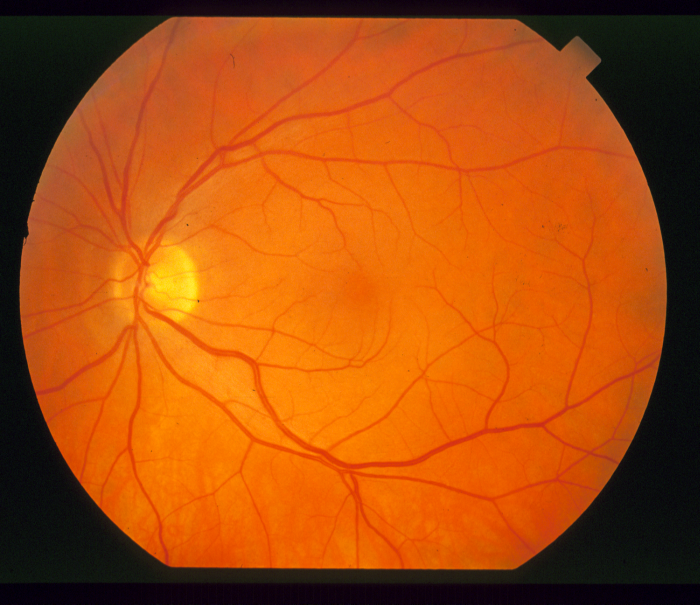}\hfill
    \includegraphics[width=0.36\linewidth]{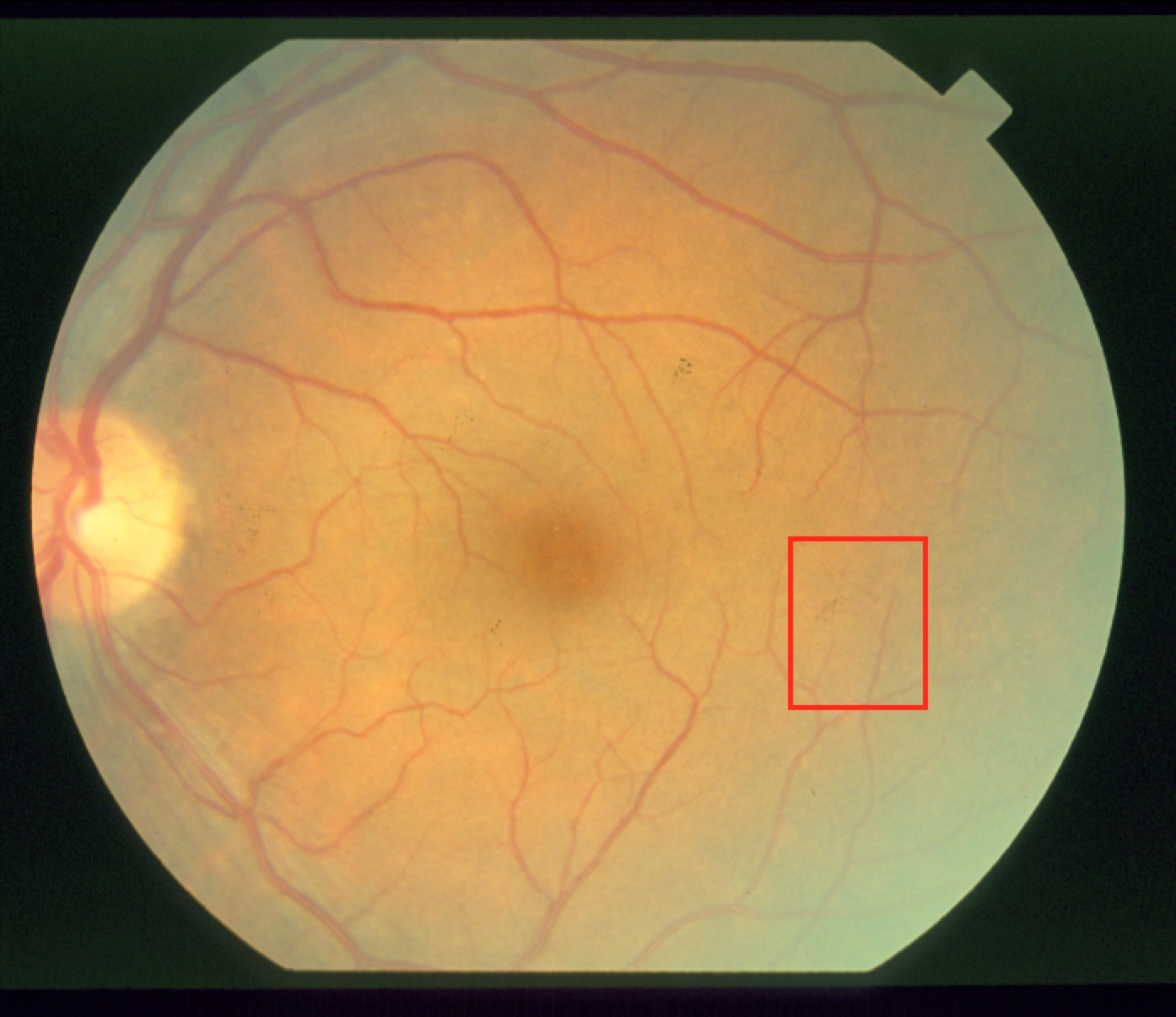}\hfill
	\cfbox{red}{\includegraphics[width=0.25\linewidth]{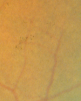}}\\[2pt]
	\includegraphics[width=0.36\linewidth]{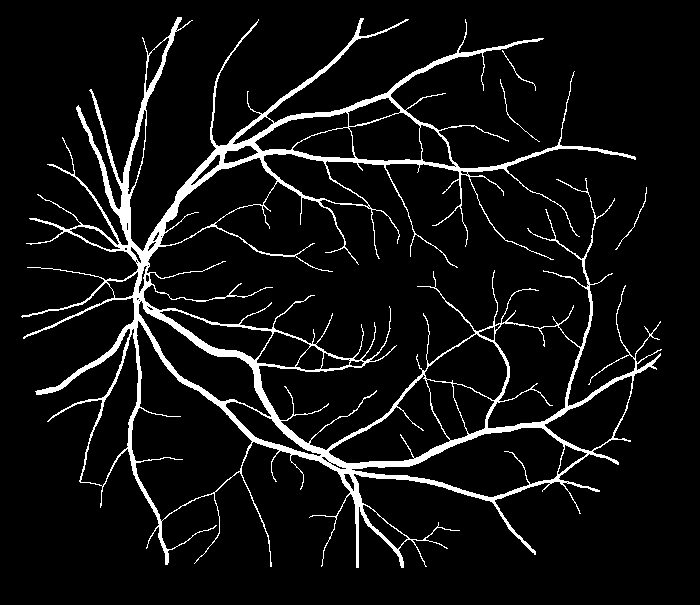}\hfill 
    \includegraphics[width=0.36\linewidth]{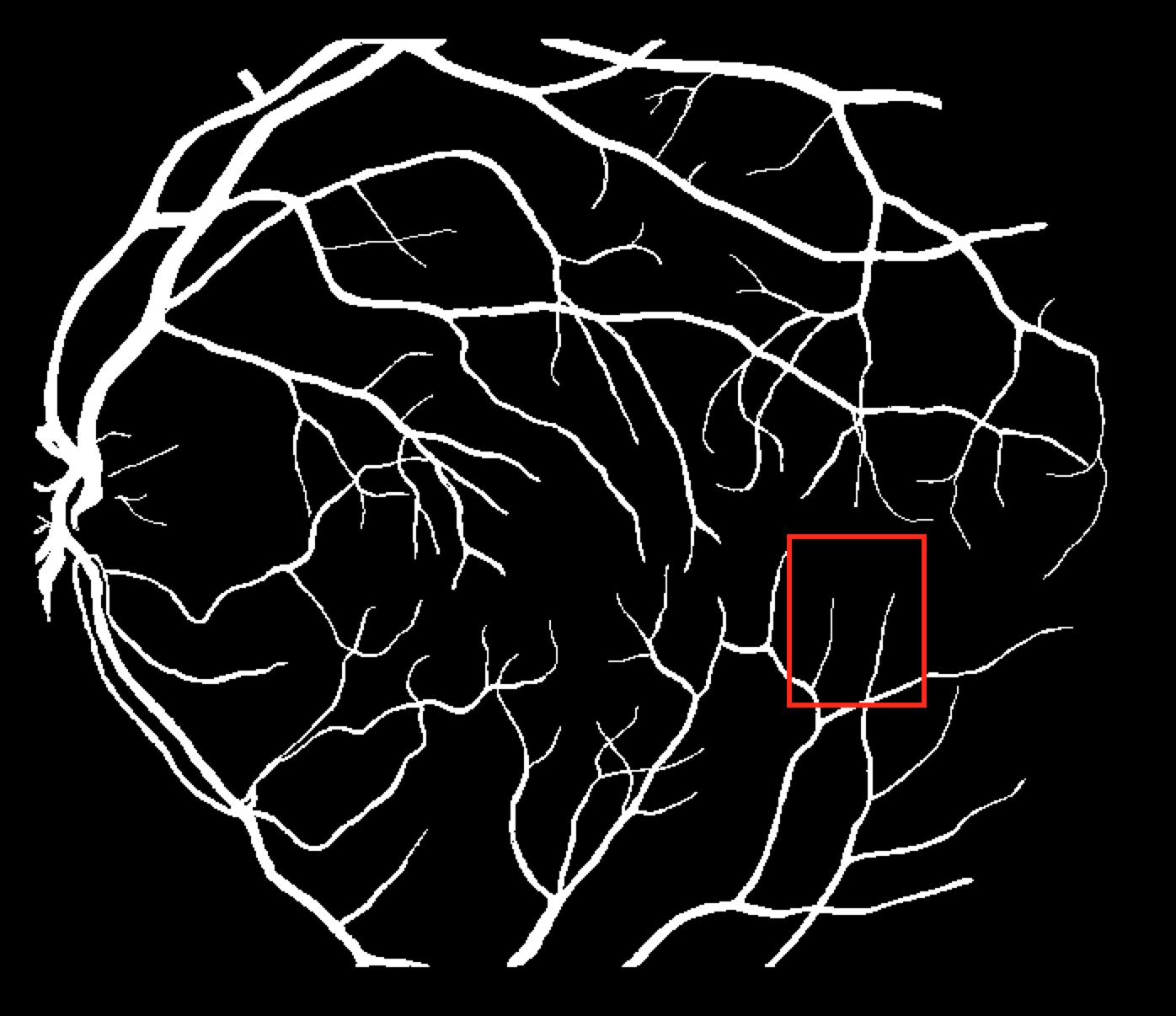}\hfill 
	\cfbox{red}{\includegraphics[width=0.25\linewidth]{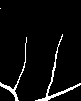}}\\[2pt]
	\includegraphics[width=0.36\linewidth]{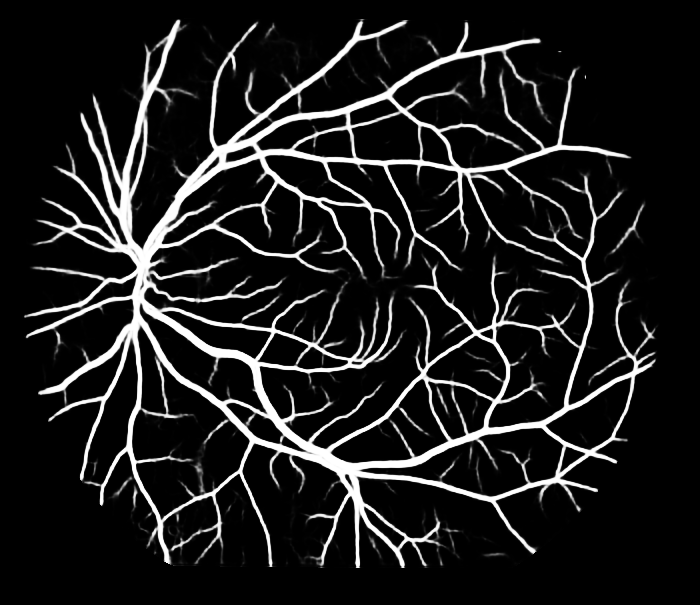}\hfill
    \includegraphics[width=0.36\linewidth]{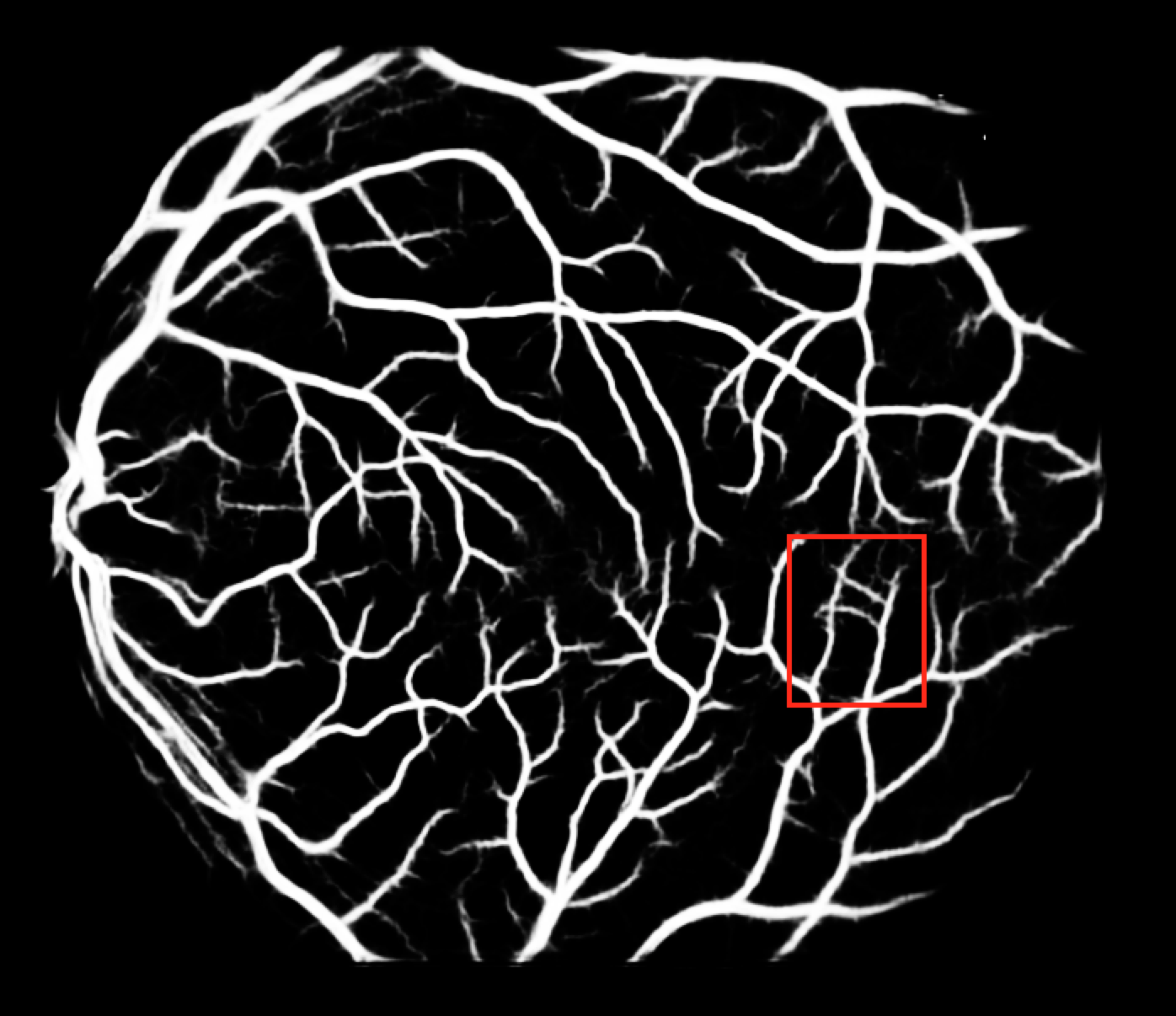}\hfill
	\cfbox{red}{\includegraphics[width=0.25\linewidth]{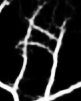}}\\[-2.5mm]
    \cfbox{white}{\rule{0mm}{1pt}}
\end{minipage}
\hfill
\begin{minipage}[b]{0.5\linewidth}
\centering
\scalebox{0.63}{%
\begin{tikzpicture}[/pgfplots/width=1.7\linewidth, /pgfplots/height=1.7\linewidth]
    \begin{axis}[% Axis labels
                 ymin=0.3,ymax=1,xmin=0.3,xmax=1,
    			 % Axis labels
        		 xlabel=Recall,
        		 ylabel=Precision,
         		 xlabel shift={-2pt},
        		 ylabel shift={-3pt},
         		 % General appearance
		         font=\small,
		         axis equal image=true,
		         enlargelimits=false,
		         clip=true,
		         % Grids 
        	     grid style=dotted, grid=both,
                 major grid style={white!65!black},
        		 minor grid style={white!85!black},
		 		 xtick={0,0.1,...,1.1},
        		 ytick={0,0.1,...,1.1},
         		 minor xtick={0,0.02,...,1},
		         minor ytick={0,0.02,...,1},
		         xticklabels={0,.1,.2,.3,.4,.5,.6,.7,.8,.9,1},
		         yticklabels={0,.1,.2,.3,.4,.5,.6,.7,.8,.9,1},
        		 % Legend
				 legend cell align=left,
        		 legend style={at={(0.02,0.02)},anchor=south west},
		         % Title
		         title style={yshift=-1ex,},
		         title={STARE - Region Precision Recall}]
        
    % Iso-f curves
    \foreach \f in {0.1,0.2,...,0.9}{%
       \addplot[white!50!green,line width=0.2pt,domain=(\f/(2-\f)):1,samples=200,forget plot]{(\f*x)/(2*x-\f)};
    }
    
    \addplot+[forget plot,only marks, mark=*, mark size=1,mark options={fill=red,draw=red}] table[x=Recall,y=Precision]
            {data/pr/STARE_test_fr_Human_points.txt};
            
    	% Ours
    \addplot+[black,solid,mark=none, line width=\deflinewidth,forget plot] table[x=Recall,y=Precision] {data/pr/STARE_test_fr_VGG.txt};
    \addplot+[black,solid,mark=o, mark size=1.3, mark options={solid},line width=\deflinewidth] table[x=Recall,y=Precision] {data/pr/STARE_test_fr_VGG_ods.txt};
    \addlegendentry{[\showodsf{STARE_test}{fr}{VGG}] Ours}
    
    	% HED
    \addplot+[green,solid,mark=none, line width=\deflinewidth,forget plot] table[x=Recall,y=Precision] {data/pr/STARE_test_fr_HED.txt};
    \addplot+[green,solid,mark=square, mark size=1.25, line width=\deflinewidth] table[x=Recall,y=Precision] {data/pr/STARE_test_fr_HED_ods.txt};
    \addlegendentry{[\showodsf{STARE_test}{fr}{HED}] HED~\cite{XiTu15}}
   
        % Wavelets
    \addplot+[gray,solid,mark=none, line width=\deflinewidth,forget plot] table[x=Recall,y=Precision] {data/pr/STARE_test_fr_Wavelets.txt};
    \addplot+[gray,solid,mark=square, mark size=1.25, line width=\deflinewidth] table[x=Recall,y=Precision] {data/pr/STARE_test_fr_Wavelets_ods.txt};
    \addlegendentry{[\showodsf{STARE_test}{fr}{Wavelets}] Wavelets~\cite{Soa+06}}
    
    % Line Detectors
    \addplot+[orange,solid,mark=none,line width=\deflinewidth,forget plot] table[x=Recall,y=Precision] {data/pr/STARE_test_fr_Lines.txt};
    \addplot+[orange,solid,mark=x, mark size=1.6, mark options={solid},line width=\deflinewidth] table[x=Recall,y=Precision] {data/pr/STARE_test_fr_Lines_ods.txt};
    \addlegendentry{[\showodsf{STARE_test}{fr}{Lines}] Line Detectors~\cite{RiPe07}}

    % Human
    \addplot+[only marks,blue,mark=+,mark size=2.7,ultra thick] table[x=Recall,y=Precision] {data/pr/STARE_test_fr_human_ods.txt};
    \addlegendentry{[\showodsf{STARE_test}{fr}{human}] \textit{Human}}
    
    \end{axis}
 \end{tikzpicture}}
 \vspace{-19pt}
\end{minipage}
\caption{\textbf{Vessel Segmentation on DRIVE (top) and STARE (bottom)}. \textbf{Left}: top row, eye fundus images; middle row, human expert annotations; bottom row, results obtained by our method.
\textbf{Right}: Region precision recall curves, methods in italics refer to pre-evaluated results.  The red dots indicate the performance of the second human annotator on each image of the test set.}
\label{fig:vessel}
\end{figure}
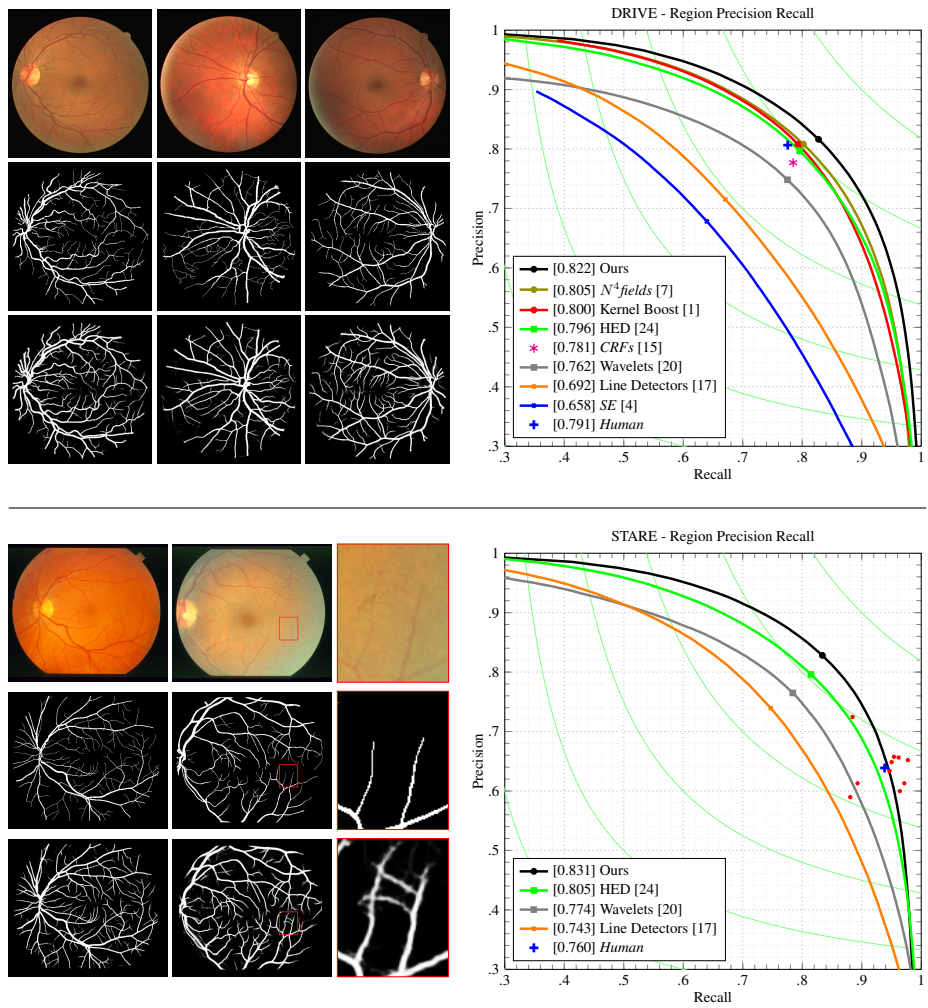

We compare the techniques by binarizing the soft-map result images at multiple 
confidence values and computing the pixel-wise precision-recall between the obtained mask and the gold standard, resulting in one curve per technique.
Figure~\ref{fig:vessel} shows both qualitative and quantitative results.
As a summary measure (in brackets in the legend) we compute the Dice coefficient or F1-measure (equivalent~\cite{Pont-Tuset2015c} to the Jaccard index) of the optimal point (marked in the lines).

The results show that DRIU performs better than all methods of comparison on both datasets, in all operating regimes. Also in both datasets, DRIU provides more consistent detections with respect to the gold standard than the second human expert (better F measure).
Interestingly, taking a closer look at some false positives of our technique (see Figure~\ref{fig:vessel} red rectangles on the bottom half), we observe that, although very weak, there are actually two vessels where DRIU signals vessels, although the human annotator did not notice them.

\paragraph{\textbf{Optic Disc Segmentation:}}
We experiment on the DRIONS-DB~\cite{DRIONS08} and RIM-ONE (r3)~\cite{RIM11} datasets (110 and 159 images, respectively).
Both contain manual segmentations of the optic disc by two expert annotators.
As for the vessels, we use the segmentations of the first annotator to train/test our algorithm.
We split into training and testing sets (60/50 and 99/60, respectively).
Given the nature of the results, apart from the region precision-recall curves, we also measure the boundary error as the mean distance between the boundary of the result and that of the ground truth.

\begin{figure}[h!]
\includegraphics[width=\linewidth]{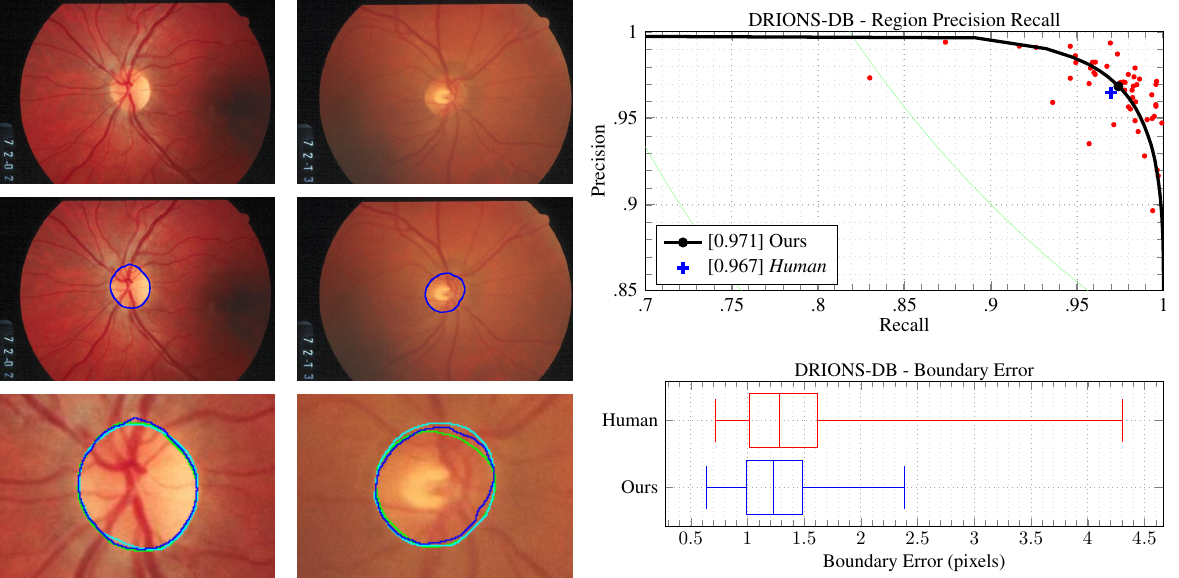}\\[2mm]
\hrule\rule{0mm}{2mm}\\
\includegraphics[width=\linewidth]{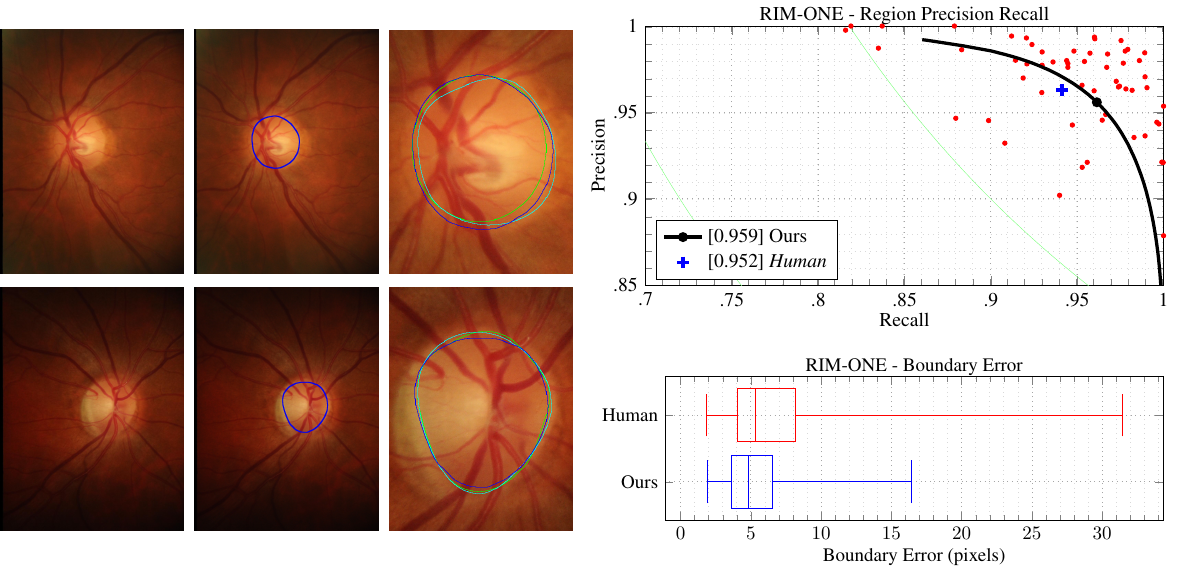}
\caption{\textbf{Optic Disc Segmentation on DRIONS-DB (top) and RIM-ONE (bottom)}. \textbf{Left}: top row, eye fundus images; middle row, our segmentation; bottom row, detail of our segmentation (blue) against the two human annotations (green and cyan). 
\textbf{Right}: Region precision-recall curves (top) and boundary error (bottom). The red dots on the curve indicate the performance of the second human annotator on each image of the test set.}
\label{fig:disc}
\end{figure}

Figure~\ref{fig:disc} shows the results of the qualitative and quantitative evaluation performed on these two datasets.
Focusing first on the region precision-recall curves, DRIU shows super-human 
performance, meaning that it presents results more coherent to the gold standard than the second human annotator.

In terms of boundary accuracy, the box-plots show the distribution of errors (limits of the 4 quartiles). In both datasets DRIU presents results that not only have a lower median error, but also show less dispersion, so more consistency.
The qualitative results corroborate that DRIU is robust and consistent, even in the more challenging and diverse scenarios of RIM-ONE database.

\section{Conclusions and Discussion}
We presented DRIU, a method for retinal vessel and optic disc segmentation that is both fast and accurate. DRIU brings the power of CNNs, which have proven groundbreaking in other fields of computer vision, to retinal image analysis by the use of a base shared CNN network and per-task specialized layers. The experimental validation in four public datasets, both qualitative and quantitative, shows that DRIU has super-human performance in these tasks\footnote{All the resources of this paper, including code and pre-trained models to reproduce the results, are available at: \url{http://www.vision.ee.ethz.ch/~cvlsegmentation/}}.

The impact of an automated solution to the problems of vessel and optic disc segmentation goes beyond assisting specialists in the initial diagnosis of eye diseases. Accurate and repeatable measurements can be an invaluable tool for monitoring their evolution. Looking forward, our technology also has the potential of changing medical practice, as it offers the possibility of carrying out robust comparative statistical analyses on large populations.

\paragraph{\textbf{Acknowledgements:}} Research funded by the EU Framework Programme for Research and Innovation - Horizon 2020 - Grant Agreement No. 645331 - EurEyeCase. We thank NVIDIA Corporation for donating the GPUs used in this project.

\bibliographystyle{splncs03}
\bibliography{paper398}

\end{document}